\documentclass{article}
\usepackage{microtype}      
\usepackage{jtdh}
\usepackage[utf8]{inputenc} 
\usepackage[english]{babel} 
\usepackage{url}   
\AtBeginDocument{\urlstyle{APACsame}}
\usepackage[hang,flushmargin]{footmisc} 
\usepackage[colorlinks = true, linkcolor = black, urlcolor  = gray, citecolor = black, anchorcolor = black]{hyperref}
\usepackage{booktabs}       
\usepackage{amsfonts}       
\usepackage{nicefrac}       
\usepackage{xcolor}
\usepackage{graphicx}
\usepackage{tabularx}
\usepackage[]{apacite}
\usepackage{enumitem}
\usepackage{amsmath}
\usepackage{adjustbox}
\usepackage{comment}
\graphicspath{{img/}}
\AtBeginDocument{%
   \renewcommand{\BRetrievedFrom}{}
}

\title{Generative Model for Less-Resourced Language with 1 billion parameters}

\author{
  Domen VREŠ,\textsuperscript{1} Martin BOŽIČ,\textsuperscript{1} Aljaž POTOČNIK,\textsuperscript{2} Tomaž MARTINČIČ,\textsuperscript{2} Marko ROBNIK-ŠIKONJA\textsuperscript{1} 
}
\institutions{
  \textsuperscript{1}University of Ljubljana, Faculty of Computer and Information Science \par
  \textsuperscript{2}XLAB d.o.o.
}

\begin{document}
\maketitle

\begin{abstract}
Large language models (LLMs) are a basic infrastructure for modern natural language processing. Many commercial and open-source LLMs exist for English, e.g., ChatGPT, Llama, Falcon, and Mistral. As these models are trained on mostly English texts, their fluency and knowledge of low-resource languages and societies are superficial. We present the development of large generative language models for a less-resourced language. GaMS 1B - Generative Model for Slovene with 1 billion parameters was created by continuing pretraining of the existing English OPT model. We developed a new tokenizer adapted to Slovene, Croatian, and English languages and used embedding initialization methods FOCUS and WECHSEL to transfer the embeddings from the English OPT model. We evaluate our models on several classification datasets from the Slovene suite of benchmarks and generative sentence simplification task SENTA. We only used a few-shot in-context learning of our models, which are not yet instruction-tuned. For classification tasks, in this mode, the generative models lag behind the existing Slovene BERT-type models fine-tuned for specific tasks. On a sentence simplification task, the GaMS models achieve comparable or better performance than the GPT-3.5-Turbo model. 
\end{abstract}
\keywordseng{large language models, generative models, knowledge transfer, OPT model, GaMS model, language adaptation}

\section{Introduction}
\label{sec:introduction}

Large language models (LLMs), in particular generative LLMs like GPT models \cite{gpt3, gpt4}, have dramatically transformed natural language processing (NLP), advancing the understanding and generation of human language. As a result of this rapid development, new open-source decoder-type transformer LLMs such as Llama, Falcon, Mistral, and many others are released on a monthly basis. These models are trained on high-resource languages (primarily English), leaving many less-resource languages, such as Slovene, behind. In this work, we present the development of GaMS 1B (Generative Model for Slovene), the first Slovene open-source generative model with 1 billion parameters. The aim is to transfer recent advancements in language technologies from English to Slovene and, therefore, improve the technological development of Slovene. We release the model under open-source license. The creation of the model is fairly general and offers useful lessons to other less-resourced languages.

The main problem in training LLMs for Slovene is the lack of data. For example, the Llama 3 model \cite{llama3} was trained on 15 trillion tokens, while the currently available Slovene corpora contain around 11 billion tokens, a thousand times fewer. This means that training an LLM from scratch for Slovene is unfeasible. Hence, we adapt the already trained English OPT model \cite{OPT} to Slovene. To increase the amount of available training data, we also include texts from Croatian, Bosnian, and Serbian languages, which can improve the models' performance due to the language similarity. Taking an English model as a starting point raises the problem of the model's vocabulary, as the existing one is not adapted to Slovene, resulting in an inefficient tokenization of Slovene texts (i.e. considerably more tokens are generated compared to efficient tokenization). To solve this problem, we train a new tokenizer and employ embedding initialization methods WECHSEL \cite{wechsel} and FOCUS \cite{focus} to transfer the embeddings from the English model to ours with the Slovene-tailored vocabulary.

An efficient evaluation of LLMs poses an additional challenge for low-resource languages. We demonstrate that models can not be directly compared based on training/validation losses observed during generative pretraining. The main reason is different vocabularies, as distributions of their output tokens differ, impacting the cross-entropy loss computation. English models are often evaluated on benchmarks testing models' reasoning, language understanding, etc. Such benchmarks are rare in Slovene, and using machine translation on complex datasets is mostly infeasible due to contextual differences between the languages. Hence, additional effort is required to obtain and adapt such benchmarks to a new language. We evaluate our models on three benchmarks already created or adapted to Slovene: the Slovene adaptation of the SuperGLUE benchmark suite \cite{slo_super_glue}, the Slovene natural language inference dataset SI-NLI \cite{slo_nli}, and the sentence simplification dataset SENTA \cite{senta}.

The paper is organized into six sections. In Section \ref{sec:related_work}, we present related work on the development of large language models and transferring their knowledge to low-resource languages. In Section \ref{sec:data}, we present the data used for training of our GaMS model. We offer a detailed technical description of GaMS model, i.e. the training of a new tokenizer, embedding transfer methods, and training details, in Section \ref{sec:model}. In Section \ref{sec:evaluation}, we evaluate the models. We provide conclusions and directions for further work in Section \ref{sec:conclusion}.

\section{Related work}
\label{sec:related_work}

New LLMs (or model families) are released on a monthly basis, with the most notable representatives being LLaMa \cite{Llama1, Llama2, llama3}, Falcon \cite{falcon}, Phi \cite{phi}, Mistral \cite{mistral}, and Mixtral \cite{mixtral}. Most of these models were trained on mainly English texts, and those trained on more languages have seen a very small proportion of Slovene texts compared to more represented languages. Therefore, the performance of these models for Slovene can be improved with additional pretraining on Slovene texts.

To spread the benefits of LLMs to languages other than English, multilingual models were developed. BLOOM \cite{bloom}, YAYI 2 \cite{yayi}, PolyLM \cite{polylm} and XGLM \cite{xglm} were all trained on over 15 languages. However, they do not achieve the performance of state-of-the-art English models due to a lower number of parameters or smaller training data size. Additionally, Slovene is not included in the supported languages or is included in such a minority that the models do not work well for Slovene.

Recently, some English models were adapted for specific languages. Most notable examples are GPT-SW3 \cite{gpt-sw3} for Swedish, Chinese LLaMa \cite{chinese_llama} and Open-Chinese-LLaMA \cite{open_chinese_llama} for Chinese, and Gervasio \cite{gervasio} for Portuguese. However, these models were either trained from scratch (GPT-SW3), did not use embedding transfer methods after vocabulary expansion (Chinese LLaMA and Open-Chinese-LLaMa), or were just instruction tuned for the target language (Gervasio).

Slovene is not without LLMs, though. However, existing works focused on encoder-type models, such as CroSloEngual BERT \cite{cse_bert} and SloBERTa \cite{sloberta}, or encoder-decoder-type models, such as SloT5 \cite{slot5}. The only working open-source decoder-type model for Slovene we are aware of is GPT-sl-base \cite{gpt_sl}, which has only 100 million parameters and was trained on only 5 billion unique tokens and is therefore not comparable to the proposed model.

\section{Pretraining data}
\label{sec:data}

LLMs require huge training sets. We use existing Slovene corpora for additional pretraining of our model. Our training corpora covers different types of text, such as news articles (Trendi \cite{trendi} - up to and including September 2023), academic works (KAS \cite{kas}), web crawls (mC4 \cite{mc4}, MaCoCu \cite{macocu_sl}, CC100 \cite{cc100}), and a mixture of them (Metafida \cite{metafida}). These corpora collectively contain around 10 B tokens, while Hoffman scaling laws \cite{hoffman_scaling} suggest 20 B tokens as a suitable quantity for 1 B model. Note that pretraining of the recent Llama 3 model \cite{llama3} used even more tokens than these scaling laws suggest resulting in still better model performance. For these two reasons, we also include Croatian, Bosnian, and Serbian texts to increase our training data. We hypothesize that using these languages should improve the model's performance due to their similarity to Slovene. This was also shown in previous works, such as CroSloEngual BERT \cite{cse_bert}. Additionally, we use English Wikipedia \cite{wikipedia} and CC-News \cite{cc_news} to prevent the model's forgetting of English. The used corpora and their properties are shown in Table \ref{tab:data}.

We performed an additional cleaning of the KAS corpus, containing some unwanted artifacts due to the scanning of PDF documents. We cleaned these artifacts using the following heuristics. We define a set of problematic characters (Non-ASCII characters except Slovene characters (č, ž, š) and characters of other alphabets, such as Chinese, Greek, Cyrillic, etc.). We consider an unwanted artifact a sequence of tokens (texts are tokenized using NLTK \cite{nltk} tokenizer) with a combined length of at least 5 characters that contain only problematic characters. We remove these sequences. We did not clean other corpora, as they were already thoroughly cleaned.

We performed near deduplication on Slovene corpora using the Onion tool \cite{onion}. Similarly to \citeA{cse_bert}, we use 9-grams with a duplicate content threshold of 0.9. The statistics, shown in Table \ref{tab:data}, are computed on cleaned and deduplicated corpora.

\begin{table}[tb]
\caption{Corpora used for additional pretraining of GaMS 1B model. CBS stands for a combination of Croatian, Bosnian, and Serbian languages. The "OPT tokenizer" column shows the number of resulting tokens when the texts are tokenized with the original OPT tokenizer, while the "Slovene tokenizer" shows the number of tokens when the texts are tokenized with our tokenizer, described in Section \ref{sec:vocab}.}
\label{tab:data}
\begin{tabularx}{\textwidth}{X|X|r|r}
\toprule
\textit{Corpus} & \textit{Language} & \textit{\# tokens (OPT tokenizer)} & \textit{\# tokens (Slovene tokenizer)} \\
\midrule
Metafida & Slovene & 6.59 B & 3.35 B \\
KAS & Slovene & 3.61 B & 1.66 B \\
Trendi & Slovene & 1.4 B & 0.68 B \\
mC4 & Slovene & 5.5 B & 2.88 B \\
MaCoCu & Slovene & 4.68 B & 2.34 B \\
CC100 & Slovene & 0.54 B & 0.29 B \\
Rižnica & Croatian & 0.21 B & 0.11 B \\
HrNews & Croatian & 4.16 B & 2.14 B \\
MaCoCu & CBS & 15.65 B & 8.63 B \\
Wikipedia & English & 4.7 B & 5.61 B \\
CC-News & English & 0.4 B & 0.46 B \\
\midrule
\textbf{Total} & / & \textbf{47.44 B} & \textbf{28.13 B} \\
\bottomrule
\end{tabularx}
\end{table}
\vskip 1cm

\section{Additional pretraining process}
\label{sec:model}

Due to the lack of Slovene data, we cannot train the model from scratch and have to use an already pretrained English model as a starting point. Most generative decoder-based LLMs have more than 1 B parameters as their performance scales with the number of parameters. There were not many open-source 1 B models available at the time we started with our work; e.g., Microsoft's Phi-1.5 \cite{phi} was not yet published under an open-source license. Two possible choices were Falcon \cite{falcon} and OPT \cite{OPT}. While Falcon models generally perform better on LLM leaderboards, such as \citeA{open-llm-leaderboard}, the 1 B version was trained only on a 350 B tokens subset of the RefinedWeb dataset \cite{falcon}. For comparison, the larger versions were trained on the whole RefinedWeb (1.19 T tokens) and some other corpora, resulting in a training dataset of around 1.5 T tokens. Even the authors of the 1 B Falcon model advise treating this model only as a research artifact. By manually testing the 1 B Falcon model on some prompts, we found out that the model tends to repeat itself (even with sampling), generates longer outputs than necessary, and outputs meaningless sentences on a regular basis. An example of such output is shown in Table \ref{tab:falcon}. When testing the 1.3 B version of the OPT model in a similar way, it made a better impression, and we chose it as our starting model.

\begin{table}[tb]
\caption{An example of an 1 B Falcon model output. The output shows clear signs of repetition and a tendency to generate too-long texts. The generation was interrupted, when the maximal number of allowed tokens was reached.}
\label{tab:falcon}
\begin{tabularx}{\textwidth}{X}
\toprule
\textbf{Prompt:} This is the initial run of the Falcon model. My first message to the world is: \\
\midrule
\textbf{Output:} I’ve done my research. I’m not a fan of the Falcon and I think the Falcon is a bad idea. I’ve spent some time on this site looking at the Falcon and the Falcon is not a very good idea. I’ve read about it and I have read a lot about it and the Falcon is not a very good idea. It is a very expensive, very inefficient, and very bad idea. It is an expensive, inefficient, and terrible idea.

The reason I am writing this post is that I want people to see what I’m talking about and to see why I think the Falcon is a bad idea. I am not a huge fan of the Falcon and I think it’s a horrible idea. The Falcon is a terrible idea. The Falcon is a terrible idea. It’s an incredibly bad idea. It \\
\bottomrule
\end{tabularx}
\end{table}
\vskip 1 cm

OPT follows the GPT-3 architecture \cite{gpt3}. The 1.3 B model has $24$ layers with $32$ attention heads. Its hidden (embedding) dimension is $2048$, it uses Pre-LayerNorm \cite{pre_ln}, ReLU activation function, absolute learned positional embeddings,  and the encoder sequence length (context length) is $2048$. It offsets the positional embeddings by $2$ (instead of starting with $0$, it starts with $2$), appends EOS token at the beginning of the sequence, and its vocabulary size is $50.272$. 

We additionally pretrain two versions of this model, one with the original OPT vocabulary and the other with Slovene vocabulary (see Section \ref{sec:vocab}). We refer to the versions with the original OPT vocabulary as OPT\_GaMS models and to versions with Slovene vocabulary as GaMS models for the rest of this paper.

\subsection{Building Slovene vocabulary of the model}
\label{sec:vocab}

We train the tokenizer for the new vocabulary using the CC100, KAS, Metafida, and HrNews \cite{cro_news} corpora. We initially trained six different tokenizers, primarily differing in size. Our aim for the tokenizer is to be efficient on both English and Slovene texts. For vocabulary evaluation, we utilize the OpenSubtitles \cite{lison2016opensubtitles2016} dataset, which includes Slovene and English subtitles, totaling around 19 million 
aligned lines in these two languages.

To train the tokenizer, we utilize the SentencePiece library \cite{kudo2018sentencepiece} with the Byte Pair Encoding (BPE) \cite{BPEEncoding} segmentation algorithm. We create a SentencePiece tokenizer model with a specified vocabulary size and include special tokens such as `<s>` (beginning of sequence), `</s>` (end of sequence), `<pad>` (padding token), and `<unk>` (unknown token).

\begin{figure}[htb]
    \caption{The evaluation of different vocabulary sizes tested on the Slovene Subtitles dataset (upper two graphs) and the English Subtitles dataset (lower two graphs).}
    \centering
    \includegraphics[width=\textwidth]{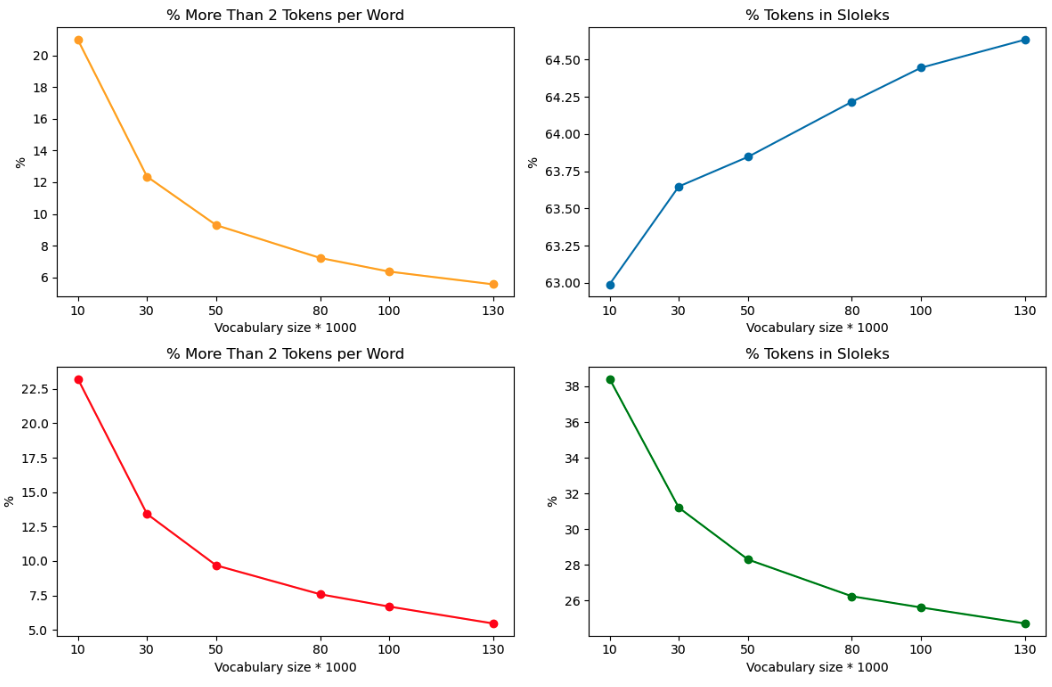}
    \label{fig:vocabulary_evaluation}
\end{figure}

We evaluate the tokenizer using three metrics. The first metric was described by \citeA{ali2023tokenizer} and measures how many words are written with two or more tokens. A good tokenizer shall keep this number relatively low. The second metric assesses how many vocabulary tokens are part of the Slovene lexical database Sloleks \cite{11356/1230}. We wish for a high value of this metric. Lastly, we create a distributional histogram displaying 10 different groups of columns, illustrating for each tokenizer the number of words written with $1$, $2$, ..., up to $10$ or more tokens. We wish for the bulk of mass in the histogram to be on the left-hand side of the histogram. We show the results of the first two metrics, evaluated on Slovene and English subtitles datasets, in Figure \ref{fig:vocabulary_evaluation}.

The results on the Slovene and English OpenSubtitles datasets show that larger vocabularies yield better results\footnote{We observe that the percentage of tokens in Sloleks increases when evaluated on the Slovene dataset and decreases when evaluated on the English dataset. This trend is favorable in both cases. Initially, we have tokenized parts of words that may be similar across both languages. As the token size increases, more complete English words, which are not in the Sloleks dictionary, appear, while more complete Slovene words, which are in Sloleks, also appear.}. However, the improvement in results slightly diminishes when the vocabulary size increases from 80,000 to 100,000 tokens. As a larger vocabulary implies more model parameters, which consequently require more data for training, more required computational resources and longer training times, we have to settle for a suitable sweet spot. We opt for a vocabulary size of 80,000 tokens as our choice for the 1 B model.

\subsection{Embedding transfer}
\label{sec:transfer}

\citeA{vocabulary_problem} recently showed that vocabulary change (or expansion) can have a negative impact on the model's performance when the new embedding matrix is initialized randomly. They performed their experiments using Chinese LLaMA \cite{chinese_llama}. As the Chinese language uses specific characters that are not well-represented in the vocabularies of English LLaMA models, the vocabulary change (or expansion) seemed a necessary step in adapting the model for Chinese. However, if vocabulary change/expansion had a negative effect for Chinese models, it should have an even more negative impact for the Slovene model. Nevertheless, the benefit of a vocabulary change is a larger context window of the model. While the number of tokens the model can process (2048 in case of OPT) is not dependent on its vocabulary, the amount of text that can be tokenized using this number of tokens is. As seen in Table \ref{tab:data}, tokenizing the same amount of Slovene text with OPT tokenizer results in twice as many tokens as tokenizing it with Slovene tokenizer. Hence, when using the Slovene tokenizer, the model can process Slovene texts that are double the size of those processed by the OPT tokenizer.

To keep the upsides of vocabulary change and mitigate its adverse effect on the model's performance, we tried to initialize the embedding matrix using WECHSEL \cite{wechsel} and FOCUS \cite{focus} initialization methods. These methods initialize the embedding matrix for a new vocabulary based on the embedding matrix of the original vocabulary. Let $T^s$ be the source tokenizer (OPT tokenizer in our case) with vocabulary $V^s$ and corresponding embedding matrix $E^s$. We have a target tokenizer $T^t$ (tokenizer from Section \ref{sec:vocab}) with vocabulary $V^t$. Our goal is to initialize the embedding matrix $E^t$. WECHSEL and FOCUS do that by computing the similarities between tokens in a common embedding space $W$. We denote the representations of $V^s$ and $V^t$ in $W$ with $W^s$ and $W^t$. Both WECHSEL and FOCUS use FastText embeddings \cite{fast_text} as $W$. We test both the original versions of these methods and our own versions, where we replace the FastText embeddings with CroSloEngual BERT embeddings \cite{cse_bert}.

We denote models obtained by using WECHSEL/FOCUS as WECHSEL/FOCUS GaMS models. Additionally, OPT uses the same weights for embedding and output layer. Hence, it makes sense to transfer the output layer as well. We denote the models, where output layer is also transfered as WECHSEL/FOCUS Tied models.

\subsubsection{The WECHSEL embeddings transfer method }
\label{sec:wechsel}

WECHSEL \cite{wechsel} obtains representations of vocabulary in source and target embeddings $W^s$ and $W^t$ by applying monolingual fastText word embeddings to $V^s$ and $V^t$ and aligning them using the Orthogonal Procrustes method \cite{orthogonal_procrustes, orthogonal_procrustes2} with bilingual dictionaries\footnote{WECHSEL code comes with already aligned embeddings, hence we did not need to align them.}. Based on this embeddings, it computes the cosine similarity $s_{x,y}$ between every pair $x\in V^t, y\in V^s$ using the following equation:
\begin{equation}
    s_{x, y} = \frac{{w_x^t}^T w_y^s}{\| w_x^t \| \cdot \| w_y^s \|},
    \label{eq:wechsel_sim}
\end{equation}
where column vectors $w_x^t$ and $w_y^s$ denote the representations of $x$ and $y$ in $W^t$ and $W^s$.

The target embeddings in $E^t$ are initialized as a convex combination of embeddings in $E^s$. Let $\mathcal{J}_x \subset V^s$ denote the set of $k$ nearest neighbors of $x \in V^t$ based on $s_{x, y}$ ($k$ is the hyperparameter of the method). The embedding $e_x^t \in E^t$ is then computed using the softmax function:
\begin{equation}
    e_x^t = \frac{\sum_{y \in \mathcal{J}_x}\exp(s_{x,y} / \tau) \cdot e_y^s}{\sum_{y' \in \mathcal{J}_x}\exp(s_{x,y'} / \tau)},
\end{equation}
where $e_y^s$ denotes the embedding of $y \in V^s$ in $E^s$ and $\tau$ denotes the temperature hyperparameter. We use $k=10$ and $\tau=0.1$ (these are default WECHSEL values) in our models.

\subsubsection{The FOCUS embeddings transfer method}
\label{sec:focus}

The FOCUS embeddings transfer method \cite{focus} initializes the target embeddings based on tokens that appear both in $V^s$ and $V^t$ (overlap). Let $O = V^s \cap V^t = \{o_1, o_2, ..., o_n\}$. The target embeddings of tokens in $O$ are the same as their source embeddings:
\begin{equation}
    \forall o \in O: e_o^t = e_o^s.
\end{equation}
The set of non-overlapping (additional) target tokens is defined as $A = V^t \setminus O$. The embeddings $e_a^t$ are computed based on similarities between tokens from $A$ and $O$. Hence, FOCUS does not need $W^s$ but needs only $W^t$, which is obtained by FastText. The difference between FOCUS and WECHSEL is that WECHSEL uses pretrained FastText, and FOCUS trains it on unlabeled data in the target language. Based on $W^t$, similarity $s_{a,o}$ is computed using Equation \ref{eq:wechsel_sim} for every pair $a\in A, o\in O$. For every $a \in A$, FOCUS defines the similarity score vector as:
\begin{equation}
    c_a = [s_{a, o_1}, s_{a, o_2}, ..., s_{a, o_n}].
\end{equation}
Based on $c_a$, the vector of weights $w_a$ is computed using sparsemax function \cite{sparsemax}:
\begin{equation}
    w_a = \text{sparsemax}(c_a).
\end{equation}
The target embedding $e_a^t \in E^t$ for an additional token $a \in A$ is then computed as:
\begin{equation}
    e_a^t = \sum_{o \in O} w_{a, o} \cdot e_o^s.
\end{equation}

We train the FastText model used with FOCUS on the same corpora as the tokenizer from Section \ref{sec:vocab}. We train the FastText model for $3$ epochs and include every token that occurs more than $10$ times in the training dataset. The dimension of token vectors is set to $768$.

\subsubsection{Using CroSloEngual BERT embeddings}
\label{sec:cse_embeddings}

Croatian, Slovene, and English languages, which are part of our vocabulary, are also used in the CroSloEngual BERT model (CSE BERT). Hence, we try to upgrade WECHSEL and FOCUS by using the embedding matrix of CSE BERT as a common embedding space $W$. The reasoning is that CSE BERT embeddings of similar English, Slovene, and Croatian tokens shall be aligned since they are modeled by the same model. As CSE BERT has shown some promising results on SloBench classification tasks \cite{slobench}, it should have good internal language knowledge. We expect that our approach will benefit the WECHSEL method more than FOCUS, as WECHSEL's bilingual alignment is not suitable for multi-lingual models such as ours. Even for mono-lingual models, we suspect that the linear alignment is the weakest point of WECHSEL, and our approach should address that. We refer to the models that are trained using CSE BERT embeddings as $W$ as FOCUS/WECHSEL CSE models.

We use the following approach to embed the tokens from $V^s$ and $V^t$ using CSE BERT. Let $v \in V^s \cup V^t$ be the token we want to embed. First, we tokenize it with the CSE BERT's tokenizer. We denote this tokenization with $t_v^{CSE}$. Since CSE BERT vocabulary is not the same as $V^s$ and $V^t$, $v$ is tokenized using $k \geq 1$ tokens:
\begin{equation}
    t_v^{CSE} = [t_{v, 1}^{CSE}, ..., t_{v, k}^{CSE}].
\end{equation}
Let $e_{v, i}^{CSE}, 1 \leq i \leq k$ denote the product of token $t_{v, i}^{CSE}$ with embedding matrix $E^{CSE}$ of CSE BERT (the CSE BERT embedding of token $t_{v, i}^{CSE}$). We define the common space embedding $w_v \in W$ for $v$ as:
\begin{equation}
    w_v = \frac{1}{k}\sum_{i=1}^k e_{v, i}^{CSE}.
\end{equation}

\subsection{Training the 1B models}
\label{sec:training}

We train our models on the Slovene HPC Vega computer (60 GPU nodes, each containing 4 NVIDIA A100 GPUs with 40 GB of RAM). We use the NVidia NeMo toolkit (version 1.22, container 23.10) for training, enabling efficient parallelization over multiple nodes on the model and data levels. As NeMo does not support positional embedding offset and ReLU activation, we forked the NeMo repository\footnote{\url{https://github.com/SloLama/NeMo}} and added the support for the OPT models.

We train our models on 16 nodes, using tensor parallel rank 4, enabling one instance of the model to be located on a single node, which is faster than having the model split over multiple nodes. We use a batch size of 1024, which equals around 2 million tokens (batch size in tokens is obtained by multiplying batch size with the context length of the model). Given our data, this results in 22,000 training steps for the OPT\_GaMS model and 13,400 training steps for the GaMS models. We use fused Adam optimizer with $\beta_1 = 0.9$ and $\beta_2 = 0.95$. We use a cosine learning rate scheduler with minimal learning rate $\eta_{min} = 2\cdot 10^{-5}$. The learning rate is first linearly increased from $0$ to $\eta_{max} = 10\cdot\eta_{min} = 2\cdot 10^{-4}$ during warmup steps and then decayed using cosine function to $\eta_{min}$, being equal to $\eta_{min}$ during the final constant steps. We use the following warmup and constant steps:
\begin{itemize}
    \item OPT\_GaMS: 1000 warmup steps, 1000 constant steps;
    \item GaMS: 2000 warmup steps, 500 constant steps.
\end{itemize}
When training the FOCUS/WECHSEL GaMS models, we freeze the inner parameters of the model for the first 1500 steps. During these steps, we train only the embedding and the output layer. This helps to avoid the catastrophic forgetting of the model, which can happen due to vocabulary change. We use 0.05\% of our data as a validation set. Even though this percentage seems small, it still results in around 15 or 24 million (depending on tokenizer) validation tokens, which should be enough to detect potential overfitting. Additionally, we can not afford large validation sets due to low amount of training data.

As \citeA{muennighoff2023scaling} showed, it might help to repeat the data when dealing with constrained data, we train the model for multiple epochs. We train the WECHSEL CSE GaMS model with both embedding and output layer transferred from the original OPT model (this is the best performing GaMS model on a single epoch according to validation loss) for 4 epochs. Additionally, we freeze the model's hidden layers (only the output and embedding layers are trained) for the entire first epoch. We train the whole model for the next 3 epochs. With a multi-epoch scenario, we set the LR scheduler's warmup steps to 10,000 and constant steps to 5,000.

Inspired by \citeA{phi}, we test training OPT\_GaMS model (we choose OPT\_GaMS instead of GaMS as GaMS seems to require more data due to a vocabulary change) only on "higher quality" data. We define higher quality data to be all data except web crawls; the selection includes news articles, literature, academic works, etc., and represents diverse, informative, and well-written texts. We use the following corpora: Metafida, KAS, Trendi, Rižnica, HrNews, Wikipedia, and CC-News. This results in around \textbf{21 B} tokens, encoded with OPT tokenizer. We train the model for 10,050 steps and set the LR scheduler's warmup and constant steps to 1,000 and 500, respectively. We refer to this model as OPT\_GaMS Quality Data.

\begin{figure}[!htb]
    \caption{The training (top) and validation (bottom) cross-entropy losses observed during the training. Note that the losses of OPT\_GaMS models can not be directly compared to the losses of GaMS models due to differences in the distributions.}
    \centering
    \includegraphics[width=\textwidth]{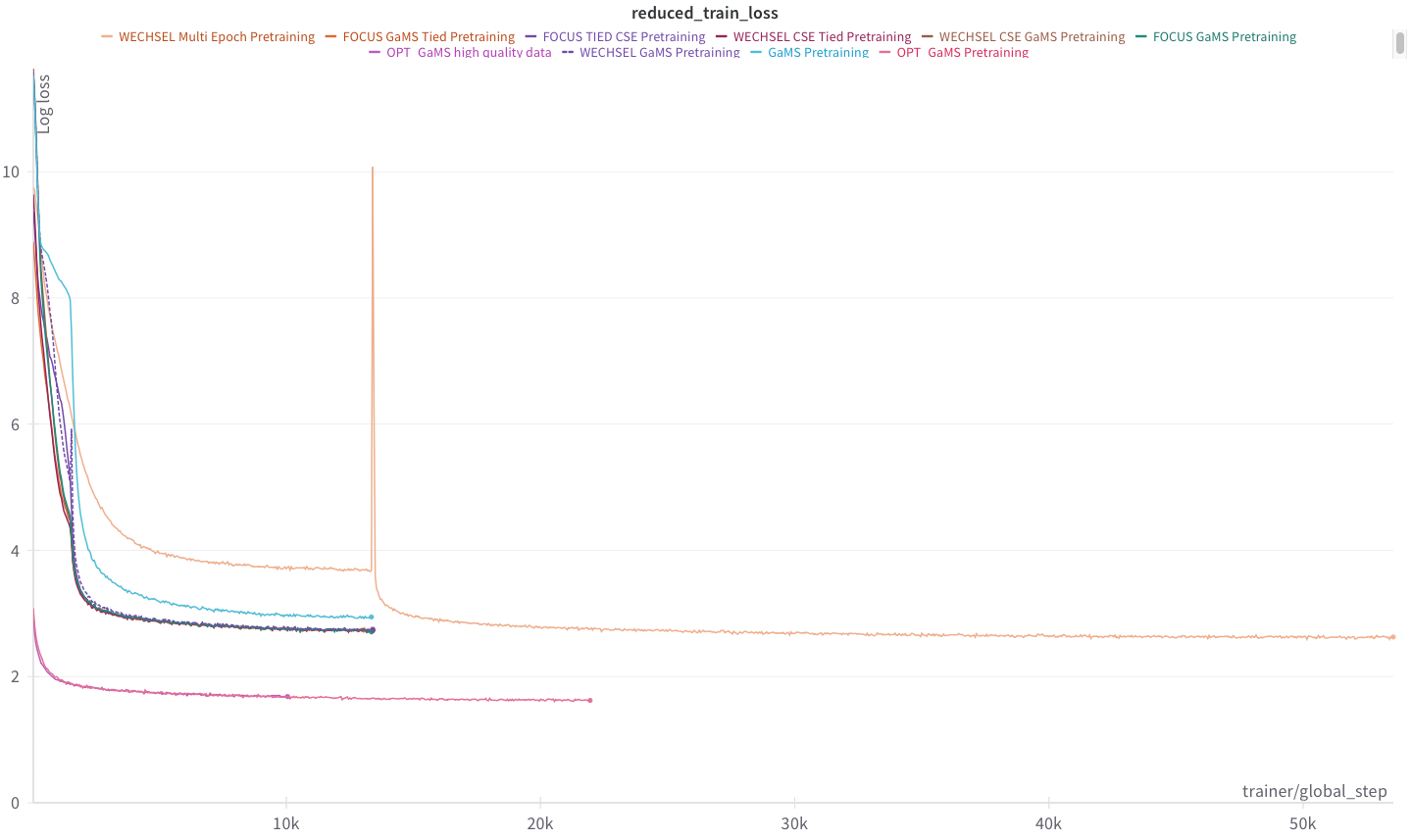}
    \includegraphics[width=\textwidth]{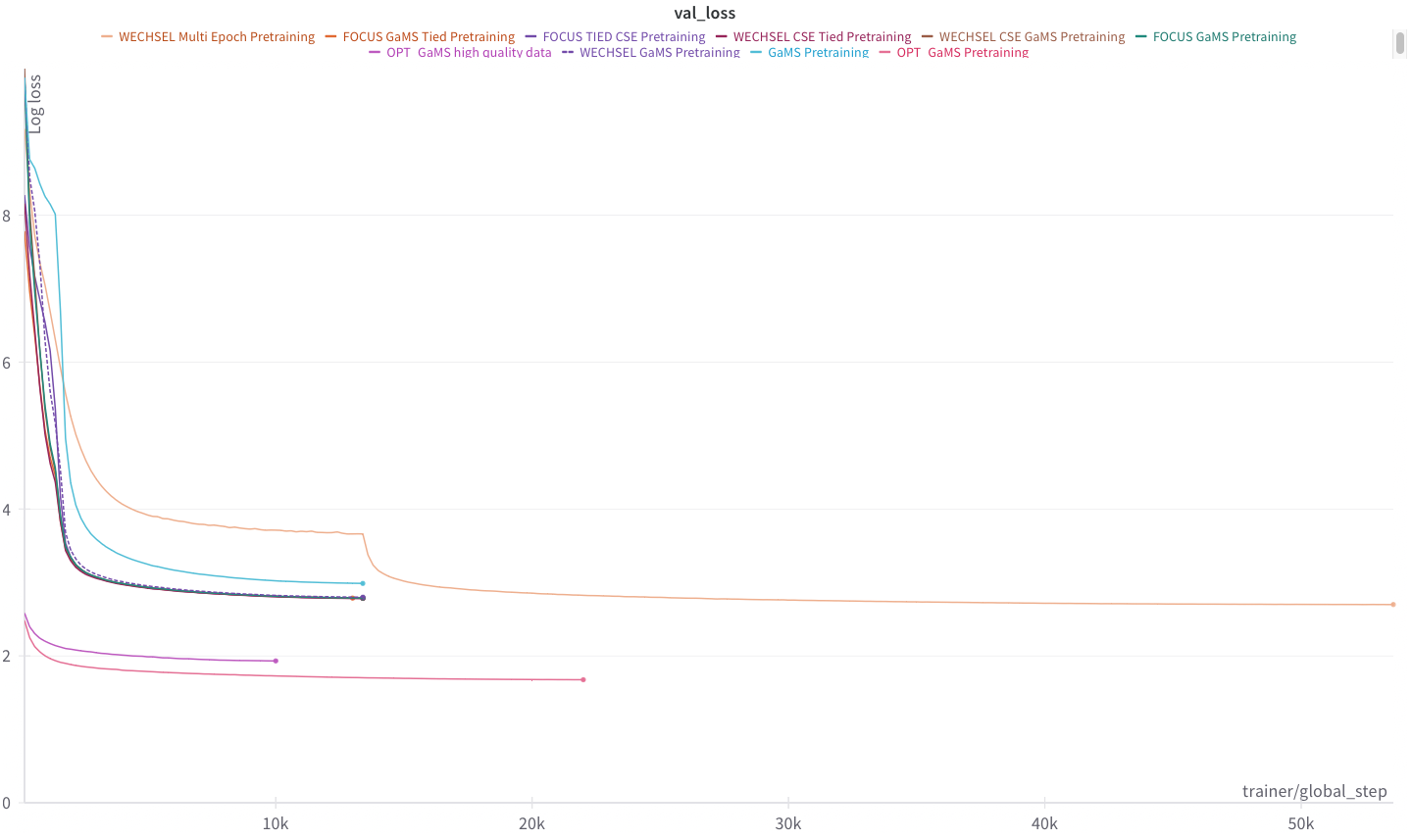}
    \label{fig:train_losses}
\end{figure}

The training and validation cross-entropy losses observed during the training are shown in Figure \ref{fig:train_losses}. The plots were obtained using Weights \& Biases platform\footnote{\url{https://wandb.ai/site}}. While GaMS losses seem to be much larger than OPT\_GaMS losses, the losses of these two model groups cannot be directly compared due to different vocabularies. Note that the loss is computed on different distributions (even though the training data is the same, it is tokenized into different tokens - even the ratios between languages are different as OPT tokenizer uses more tokens on average to tokenize Slovene words than Slovene tokenizers). To avoid unfair comparisons, we compare the losses of GaMS models. It is evident that FOCUS and WECHSEL improve the model performance compared to random initialization of the embedding matrix. While different transfer approaches behave differently in the early stages of the training, their losses all converge to a similar value (validation losses differ by less than \textbf{0.02}, showing no significant difference in the performance of these methods). Although Figure \ref{fig:train_losses} does not show this clearly, using multiple epochs actually reduces the validation loss (the final validation loss of multi-epoch model is \textbf{2.699}, while the final validation loss of its single-epoch counterpart is \textbf{2.781}). Furtherher, training the OPT\_GaMS model only on "higher quality" data does not improve its performance.

\section{Evaluation}
\label{sec:evaluation}

LLMs are commonly benchmarked for knowledge, reasoning, safety, natural language understanding, etc. The commonly used benchmarking suites for LLM evaluation in English are GLUE \cite{glue}, SuperGLUE \cite{super_glue}, BIGBench \cite{big_bench}, Massive Multitask Language Understanding (MMLU) \cite{mmlu}, etc. The benchmarks for Slovene are very limited, as due to the complexity of most LLM benchmarks, obtaining them via pure machine translation is not a viable solution. Some SuperGLUE tasks are unsuitable even for human translation due to contextual differences between languages (such as the Word in Context task) and have to be rewritten for Slovene. Besides classification tasks contained in the Slovene SuperGLUE \cite{slo_super_glue} benchmarking suite, we used two more datasets: a natural language inference classification dataset SI-NLI \cite{slo_nli}, which is already part of SloBench \cite{slobench}, and sentence simplification task SENTA \cite{senta} that tests text generation abilities of LLMs.

In our evaluation scenario, all models are evaluated using in-context learning, with few-shot prompts (models are not fine-tuned on given tasks but shown a few solved examples in the prompt). The in-context examples are randomly sampled from the training set (each test instance is given different examples). None of the models, apart from OPT\_GaMS INZ, are instruction-tuned. The OPT\_GaMS INZ model is LoRA \cite{lora} tuned on the QA dataset that was provided to us by Inštitut za novejšo zgodovino (INZ). The dataset consists of approximately 7,000 questions and answers and is not suitable for general-purpose instruction tuning, as it contains only one task. However, this fine-tuning helps with the evaluation of question-answering tasks, as it helps the model to generate the answer in the correct form. All models are evaluated using greedy sampling during the generation phase, i.e. the most probable token, according to the model, is always selected as the next generated token.

\subsection{Classification tasks}
\label{sec:classification_eval}

The number of few-shot examples and number of test set instances for each dataset from the Slovene SuperGLUE suite and the SI-NLI dataset are shown in Table \ref{tab:sbench_stats}. The number of few-shot examples in prompt ($k$) is determined based on the models' performances on the validation set. The number of test set instances is quite low for BoolQ (30) and RTE (29) because only human-translated examples are used for the evaluation.

\begin{table}[htb]
\caption{The number of test examples and the number of in-context examples in prompts ($k$) per data set in SupeGLUE tasks and SI-NLI.}
\label{tab:sbench_stats}
\centering
\begin{tabular}{l|cc}
\toprule
\textit{Task} & $k$ & \textit{\# test examples} \\
\midrule
BoolQ & 3 & 30 \\
CB & 5 & 250  \\
COPA & 5 & 500 \\
MultiRC & 2 & 333 \\
RTE & 3 & 29 \\
WSC & 4 & 146 \\
SI-NLI & 5 & 998 \\
\bottomrule
\end{tabular}
\end{table}
\vskip 1cm

To adapt the classification tasks to generative LLMs, we wrote our own framework for the evaluation of generative models, where we specify the expected form of an answer in the prompt. We observe that our 1 B models struggle to understand what output is required to complete the tasks. This is typical for models below 5 B parameters; for example, \citeA{phi} observed similar behavior for their Phi model. This behavior is not present in larger  generative models for English, especially the ones trained for instruction following. Hence, we measure the percentage of invalid predictions where a model did not generate the answer in a required form. We measure other metrics for each task only on valid predictions. The alternative would be to label the invalid predictions as wrong answers, but in this way, we cannot distinguish between invalid and wrong predictions. We also observe a high correlation between the majority label of few-shot examples and the models' output. Hence, we hypothesize that few-shot examples did not help the model to understand the tasks but only helped it with the form of the answer - more few-shot examples resulted in fewer invalid predictions.

The results for Slovene SuperGLUE tasks are shown in Tables \ref{tab:sglue_results1} and \ref{tab:sglue_results2}. Overall, the performance of the models is quite similar and there is no model that would outperform others across all tasks. The models are outperformed by the representation model CroSloEngual BERT, which was fine-tuned on these tasks. As this model has seen significantly more training instances, the comparison is not fair but the score indicates what is achievable with relatively small LLMs. The most difficult task for the models, according to the percentages of invalid predictions, is MultiRC. In this task, the model is given a text, a question, and a list of answers. The model has to return the numbers of correct answers. We could make this task easier for the models by giving them each answer separately and asking them to classify them as correct or wrong. However, as the purpose of the task is to check whether the model can select the correct answers from multiple choices, we decided to present it in this more challenging form. OPT\_GaMS INZ model produced a significantly lower percentage of invalid predictions on this task than other models, suggesting that instruction tuning should make the task less challenging.

\begin{table}[htb]
\caption{Test set results with 95 \% confidence intervals for Slovene Super GLUE tasks BoolQ, CB, and COPA. Columns Acc. represent models' accuracy, and columns Inv. pred. represent the percentage of invalid predictions for each model. Confidence intervals are computed using standard error estimation for accuracy, and using quantile bootstrap for $F_1$-score. The results for CroSloEngual BERT are copied from SloBench.}
\label{tab:sglue_results1}
\resizebox{\textwidth}{!}{
\begin{tabular}{l|cc|ccc|cc}
\toprule
& \multicolumn{2}{c|}{\textit{BoolQ}} & \multicolumn{3}{c|}{\textit{CB}} & \multicolumn{2}{c}{\textit{COPA}} \\
\textit{Model} & \textit{Acc.} & \textit{Inv. pred.} & \textit{Acc.} & \textit{$F_1$} & \textit{Inv. pred.} & \textit{Acc.} & \textit{Inv. pred.} \\
\midrule
OPT\_GaMS & 0.57 [0.38, 0.75] & 0 \% & 0.44 [0.38, 0.50] & 0.32 [0.26, 0.39] & 0 \% & 0.46 [0.42, 0.51] & 0 \% \\
GaMS & 0.50 [0.31, 0.69] & 0 \% & 0.43 [0.37, 0.50] & 0.30 [0.25, 0.33] & 1.20 \% & 0.49 [0.44, 0.54] & 17.20 \% \\
WECHSEL GaMS & 0.67 [0.49, 0.85] & 0 \% & 0.50 [0.44, 0.56] & 0.39 [0.32, 0.47] & 1.20 \% & 0.48 [0.44, 0.52] & 0.20 \% \\
FOCUS GaMS & 0.67 [0.49, 0.85] & 0 \% & 0.51 [0.45, 0.58] & 0.38 [0.31, 0.46] & 1.60 \% & 0.48 [0.43, 0.53] & 27.80 \% \\
WECHSEL CSE & 0.57 [0.38, 0.75] & 0 \% & 0.50 [0.44, 0.56] & 0.34 [0.30, 0.38] & 0.40 \% & 0.48 [0.44, 0.53] & 2.80 \% \\
WECHSEL CSE Tied & 0.47 [0.28, 0.66 & 0 \% & 0.51 [0.45, 0.57] & 0.38 [0.32, 0.46] & 2.40 \% & 0.48 [0.44, 0.53] & 0.40 \% \\
FOCUS CSE Tied & 0.50 [0.31, 0.69] & 0 \% & 0.48 [0.42, 0.54] & 0.36 [0.29, 0.44] & 0.40 \% & 0.47 [0.43, 0.51] & 3.40 \% \\
FOCUS GaMS Tied & 0.53 [0.34, 0.72] & 0 \% & 0.48 [0.42, 0.54] & 0.36 [0.29, 0.44] & 1.20 \% & 0.48 [0.43, 0.53] & 12.00 \% \\
OPT\_GaMS Quality Data & 0.60 [0.41, 0.79] & 0 \% & 0.44 [0.37, 0.50] & 0.35 [0.28, 0.43] & 0.80 \% & 0.48 [0.44, 0.52] & 0 \% \\
OPT\_GaMS INZ & 0.60 [0.41, 0.79] & 0 \% & 0.44 [0.37, 0.50] & 0.32 [0.26, 0.40] & 0 \% & 0.45 [0.41, 0.49] & 0 \% \\
WECHSEL Multi-Epoch & 0.60 [0.41, 0.79] & 0 \% & 0.51 [0.45, 0.57] & 0.38 [0.31, 0.46] & 0.80 \% & 0.46 [0.42, 0.51] & 1.20 \% \\
\midrule
CroSloEngual BERT & 0.73 & / & 0.79 & 0.74 & / & 0.57 & / \\
\bottomrule
\end{tabular}}
\end{table}
\vskip 1cm

\begin{table}[htb]
\caption{Test set results with 95 \% confidence intervals for Slovene Super GLUE tasks MultiRC, RTE and WSC. Columns Acc. represent models' accuracy, column EM represents the exact match between predictions and true labels, and columns Inv. pred. represent the percentage of invalid predictions for each model. Confidence intervals are computed using standard error estimation for accuracy and exact match, and using quantile bootstrap for $F_1$-score. The results for CroSloEngual BERT are copied from SloBench. }
\label{tab:sglue_results2}
\resizebox{\textwidth}{!}{
\begin{tabular}{l|ccc|cc|cc}
\toprule
& \multicolumn{3}{c|}{\textit{MultiRC}} & \multicolumn{2}{c|}{\textit{RTE}} & \multicolumn{2}{c}{\textit{WSC}} \\
\textit{Model} & \textit{EM} & \textit{$F_1$} & \textit{Inv. pred.} & \textit{Acc.} & \textit{Inv. pred.} & \textit{Acc.} & \textit{Inv. pred.} \\
\midrule
OPT\_GaMS & 0.15 [0.02, 0.28] & 0.43 [0.32, 0.54] & 90.09 \% & 0.41 [0.22, 0.60] & 0 \% & 0.51 [0.43, 0.60] & 0 \% \\
GaMS & 0.03 [-0.03, 0.09] & 0.16 [0.12, 0.20] & 89.49 \% & 0.43 [0.23, 0.62] & 3.45 \% & 0.42 [0.34, 0.50] & 0 \% \\
WECHSEL GaMS & 0.15 [0.03, 0.26] & 0.37 [0.30, 0.43] & 87.69 \% & 0.43 [0.23, 0.62] & 3.45 \% & 0.47 [0.38, 0.55] & 0 \% \\
FOCUS GaMS & 0.11 [0.00, 0.21] & 0.36 [0.29, 0.44] & 88.59 \% & 0.54 [0.34, 0.73] & 3.45 \% & 0.50 [0.42, 0.58] & 0 \% \\
WECHSEL CSE & 0.06 [-0.01, 0.12] & 0.26 [0.20, 0.31] & 84.08 \% & 0.43 [0.23, 0.62] & 3.45 \% & 0.45 [0.36, 0.53] & 0 \% \\
WECHSEL CSE Tied & 0.12 [0.03, 0.21] & 0.21 [0.17, 0.25] & 84.38 \% & 0.46 [0.27, 0.66] & 3.45 \% & 0.55 [0.47, 0.63] & 0 \% \\
FOCUS CSE Tied & 0.09 [0.01, 0.17] & 0.26 [0.21, 0.31] & 83.48 \% & 0.43 [0.23, 0.62] & 3.45 \% & 0.55 [0.47, 0.64] & 0 \% \\
FOCUS GaMS Tied & 0.05 [0.02, 0.08] & 0.22 [0.20, 0.24] & 36.64 \% & 0.46 [0.27, 0.66] & 3.45 \% & 0.49 [0.41, 0.58] & 0 \% \\
OPT\_GaMS Quality Data & 0.12 [0.04, 0.19] & 0.32 [0.25, 0.39] & 79.28 \% & 0.38 [0.19, 0.57] & 0 \% & 0.47 [0.39, 0.55] & 0 \% \\
OPT\_GaMS INZ & 0.07 [0.04, 0.09] & 0.34 [0.32, 0.37] & 2.10 \% & 0.38 [0.19, 0.57] & 0 \% & 0.45 [0.36, 0.53] & 0 \% \\
WECHSEL Multi-Epoch & 0.13 [0.05, 0.21] & 0.28 [0.22, 0.33] & 79.28 \% & 0.50 [0.30, 0.70] & 3.45 \% & 0.54 [0.46, 0.62] & 0 \% \\
\midrule
CroSloEngual BERT & 0.09 & 0.52 & / & 0.66 & / & 0.61 & / \\
\bottomrule
\end{tabular}}
\end{table}
\vskip 1cm

The results for the SI-NLI dataset are shown in Table \ref{tab:nli_results}. The performance of our models is quite similar, and the confidence intervals overlap. All models return invalid predictions for approximately half of the test instances (the best-performing model with respect to that metric is WECHSEL CSE, with 44.69 \% of invalid predictions). The reason for these similarities is that all models perform poorly due to lack of task understanding. Hence, the models should be instruction-tuned first in order to spot any significant differences between them.  The models are significantly outperformed by GPT and BERT models; again the comparison is not fair as BERT models were fine-tuned on this data set and GPT-3.5-Turbo is significantly larger.

\begin{table}[tb]
\caption{Test set results with 95 \% confidence intervals for the SI-NLI dataset. Columns Inv. pred. represent the percentage of invalid predictions for each model. Confidence intervals are computed using standard error estimation for accuracy and using quantile bootstrap for $F_1$-score. The results for GPT-3.5-Turbo, SloBERTa, and CroSloEngual BERT are copied from SloBench.}
\label{tab:nli_results}
\resizebox{\textwidth}{!}{
\begin{tabular}{l|ccccc}
\toprule
\textit{Model} & \textit{Accuracy} & \textit{Entailment $F_1$} & \textit{Neutral $F_1$} & \textit{Contradiction $F_1$} & \textit{Inv. pred.} \\
\midrule
OPT\_GaMS & 0.32 [0.27, 0.36] & 0.38 [0.32, 0.45] & 0.17 [0.10, 0.24] & 0.34 [0.27, 0.40] & 51.40 \% \\
GaMS & 0.29 [0.25, 0.33] & 0.31 [0.24, 0.38] & 0.32 [0.25, 0.38] & 0.25 [0.19, 0.32] & 50.00 \% \\
WECHSEL GaMS & 0.33 [0.29, 0.37] & 0.39 [0.33, 0.45] & 0.33 [0.27, 0.39] & 0.26 [0.20, 0.32] & 44.69 \% \\
FOCUS GaMS & 0.34 [0.30, 0.38] & 0.40 [0.34, 0.46] & 0.37 [0.31, 0.44] & 0.20 [0.13, 0.26] & 49.40 \% \\
WECHSEL CSE & 0.32 [0.28, 0.36] & 0.38 [0.32, 0.43] & 0.37 [0.30, 0.43] & 0.17 [0.11, 0.24] & 47.80 \% \\
WECHSEL CSE Tied & 0.35 [0.31, 0.39] & 0.40 [0.34, 0.46] & 0.41 [0.35, 0.47] & 0.20 [0.14, 0.27] & 48.30 \% \\
FOCUS CSE Tied & 0.34 [0.30, 0.38] & 0.38 [0.32, 0.44] & 0.38 [0.32, 0.44] & 0.23 [0.17, 0.30] & 47.19 \% \\
FOCUS GaMS Tied & 0.32 [0.28, 0.36] & 0.37 [0.31, 0.43] & 0.37 [0.31, 0.44] & 0.20 [0.14, 0.26] & 47.19 \% \\
OPT\_GaMS Quality Data & 0.31 [0.27, 0.35] & 0.38 [0.32, 0.44] & 0.28 [0.22, 0.35] & 0.28 [0.22, 0.35] & 47.39 \% \\
OPT\_GaMS INZ & 0.30 [0.26, 0.35] & 0.36 [0.29, 0.42] & 0.25 [0.18, 0.32] & 0.29 [0.23, 0.36] & 53.31 \% \\
WECHSEL Multi-Epoch & 0.30 [0.26, 0.34] & 0.37 [0.31, 0.43] & 0.37 [0.31, 0.43] & 0.17 [0.11, 0.24] & 51.10 \% \\
\midrule
GPT-3.5-Turbo & 0.86 & 0.85 & 0.82 & 0.90 & / \\
SloBERTa & 0.74 & 0.76 & 0.71 & 0.64 & / \\
CroSloEngual BERT & 0.66 & 0.69 & 0.63 & 0.66 & / \\
\bottomrule
\end{tabular}}
\end{table}
\vskip 1cm

\subsection{Sentence simplification}
\label{sec:sent_simp}

The models introduced in this paper are generative. Therefore, it makes sense to evaluate them on language generation tasks. We choose sentence simplification task SENTA \cite{senta}. The model is given a sentence and asked to simplify it. Here, we observe that our models perform better than in classification tasks but there are still some problems with the task understanding, as the models sometimes return "Poenostavi naslednji stavek." (eng. "Simplify the given sentence.") as an answer in case of few-shot prompts. They return this sentence, as this is the instruction added to each example in the prompt and is consequently the most common sentence in the prompt.

We evaluate our models using different values $k$ of few-shot examples. We test values $k \in \{0, 3, 5, 10\}$. We use SARI score\footnote{\url{https://huggingface.co/spaces/evaluate-metric/sari}} as an evaluation metric. SARI score is commonly used to evaluate text simplification systems. It compares the system's output to both the input and reference output. It computes the $F_1$-score for added and preserved tokens and precision for deleted words. It is computed using the following equation:
\begin{equation}
    \text{SARI} = \frac{F_{1,add} + F_{1,keep} + P_{del}}{3},
\end{equation}
where $F_{1,add}$ and $F_{1,keep}$ represent the 4-gram $F_1$ score for add/keep operations and $P_{del}$ denotes the 4-gram precision score for delete operations. The goal is to have as high $F_1$ and precision scores as possible, meaning that higher SARI score is better.

The results are shown in Table \ref{tab:sari_results}. All models perform similarly (no significant differences between their SARI scores). Using a larger number of few-shot examples seems to improve the performance of the majority of the models (the exception here is LoRA-tuned OPT\_GaMS INZ, which works best in the 0-shot scenario). Surprisingly, our models perform similarly or better than GPT-3.5-Turbo. Our best-performing model (WECHSEL GaMS in the 10-shot scenario) also outperforms the best-performing SloT5 model that was trained on this task. However, the differences in the SARI scores are not significant. We believe that the performance of our models could improve drastically with instruction-tuning, as the models would better understand the task instruction.

\begin{table}[htb]
\caption{SARI scores with 95 \% confidence intervals on SENTA task. Confidence intervals were computed using quantile bootstrap method. Value of $k$ in columns denotes the number of shown examples in few-shot prompts. The results for GPT and T5 models are copied from \protect\citeA{senta}}
\label{tab:sari_results}
\resizebox{\textwidth}{!}{
\begin{tabular}{l|cccc}
\toprule
\textit{Model} & $k=0$ & $k=3$ & $k=5$ & $k=10$ \\
\midrule
OPT\_GaMS & 39.38 [38.63, 40.16] & 38.51 [37.60, 39.46] & 39.49 [38.67, 40.40] & 39.67 [38.80, 40.60] \\
GaMS & 39.58 [38.76, 40.47] & 38.92 [37.96, 39.86] & 37.98 [37.10, 38.90] & 39.18 [38.37, 40.06] \\
WECHSEL GaMS & 39.34 [38.59, 40.15] & 39.53 [38.55, 40.43] & \textbf{39.87 [39.01, 40.77]} & \textbf{41.62 [40.82, 42.30]} \\
FOCUS GaMS & 38.50 [37.77, 39.37] & \textbf{40.16 [39.27, 41.07]} & 39.67 [38.81, 40.57] & 41.16 [40.41, 41.89] \\
WECHSEL CSE & 39.02 [38.28, 39.83] & 39.42 [38.49, 40.35] & 39.22 [38.37, 40.05] & 40.54 [39.79, 41.26] \\
WECHSEL CSE Tied & 38.77 [37.96, 39.60] & 38.67 [37.79, 39.61] & 39.29 [38.41, 40.20] & 40.91 [40.13, 41.71] \\
FOCUS CSE Tied & 38.93 [38.15, 39.77] & 38.95 [38.02, 39.92] & 39.38 [38.54, 40.25] & 40.98 [40.15, 41.79] \\
FOCUS GaMS Tied & 38.80 [37.99, 39.67] & 40.05 [39.19, 40.97] & 39.74 [38.86, 40.64] & 41.50 [40.79, 42.20] \\
OPT\_GaMS Quality Data & 38.76 [37.97, 39.58] & 37.62 [36.72, 38.47] & 38.48 [37.64, 39.44] & 39.02 [38.14, 39.91] \\
OPT\_GaMS INZ & \textbf{40.29 [39.49, 41.10]} & 37.88 [36.99, 38.88] & 38.58 [37.72, 39.54] & 38.90 [38.00, 39.85] \\
WECHSEL Multi-Epoch & 38.80 [37.96, 39.62] & 40.06 [39.23, 40.96] & 39.80 [38.97, 40.62] & 40.99 [40.23, 41.67] \\
\midrule
GPT-3.5-Turbo & \multicolumn{4}{c}{38.76} \\
SloT5-small & \multicolumn{4}{c}{39.79} \\
mT5-small & \multicolumn{4}{c}{39.09} \\
SloT5-large & \multicolumn{4}{c}{41.01} \\
\bottomrule
\end{tabular}
}
\end{table}
\vskip 1cm

We can conclude that our 1 B Slovene models are not suitable for in-context learning of classification tasks but work well in generative tasks. Their performance on classification tasks with fine-tuning remains part of the future work.

\section{Conclusion}
\label{sec:conclusion}

In this work, we presented the new 1 B Slovene generative model GaMS\footnote{\href{https://huggingface.co/cjvt/OPT_GaMS-1B}{https://huggingface.co/cjvt/OPT\_GaMS-1B}, \href{https://huggingface.co/cjvt/GaMS-1B}{https://huggingface.co/cjvt/GaMS-1B}}, which is based on the English OPT model. The model is the first fully open-source generative language model for Slovene. Based on the analysis of different vocabulary sizes, we created a new tokenizer that was trained on Slovene, English, and Croatian texts. We tested different embedding initialization methods and showed that they reduce both training and validation loss for next token prediction compared to random initialization.

The main challenge that we face in this work is a robust evaluation of the models. Direct comparison of training/validation losses for models using different vocabularies is not sensible, as the distributions of tokens (on which the loss is computed) are different. The comparison of models on classification benchmarking tasks is inconclusive, as the models do not really understand the tasks due to their size and lack of instruction tuning. We showed that our models perform better on generative tasks like sentence simplification but we need more tasks to get reliable conclusions on models performance.

In the future work, we will develop an instruction-following dataset and instruction-tune our models. This might improve the models performance on classification tasks, as the models will understand the evaluation tasks. For classification tasks, fine-tuning of models is also sensible. Additionally, we plan to train and release a larger model, where the differences between embedding initialization methods should be more significant.

\acknowledgmenteng{
The work was supported by the Slovenian Research and Innovation Agency (ARIS) research project PoVeJMo (Adaptive Natural Language Processing with the Help of Large Language Models), core research programme P6-0411, and projects J7-3159, and L2-50070. 

\bibliographystyle{apacite}
\urlstyle{same}
\bibliography{references-eng}

\newpage
\titleeng{Generativni model z milijardo parametrov za jezik z manj viri}
\begin{abstracteng}
Veliki jezikovni modeli so osnovna infrastruktura za sodobno obdelavo naravnega jezika. Za angleščino obstajajo številni komercialni in odprtokodni modeli, na primer ChatGPT, Llama, Falcon in Mistral. Ker so ti modeli učeni večinoma na angleških besedilih, sta njihovo znanje in poznavanje jezikov ter družb z manj viri površna. Predstavljamo razvoj novega generativnega velikega jezikovnega modela za jezik z malo viri. Za slovenski model, imenovan GaMS 1B (Generativni Model za Sloveščino), z 1 milijardo parametrov smo razvili nov tokenizator, prilagojen slovenščini, hrvaščini in angleščini, ter uporabili metodi inicializacije vektorskih vložitev FOCUS in WECHSEL za prenos vložitev iz obstoječega angleškega modela OPT. Zgrajene modele smo ovrednotili na slovenski zbirki klasifikacijskih učnih množic in na generativni nalogi poenostavljanja stavkov SENTA. Pri evalvaciji smo uporabili le učenje v kontekstu z nekaj učnimi primeri ter modele, ki še niso prilagojeni za sledenje navodilom. Pri takih nastavitvah so na klasifikacijskih nalogah zgrajeni generativni modeli zaostali za obstoječimi slovenskimi modeli tipa BERT, ki so bili prilagojeni za dane naloge. Pri nalogi poenostavljanja stavkov modeli GaMS dosegajo primerljive ali boljše rezultate kot model GPT-3.5-Turbo. 

\end{abstracteng}

\keywordseng{veliki jezikovni modeli, generativni modeli, prenos znanja, OPT model, GaMS model, jezikovno prilagajanje}

\end{document}